\documentclass[lettersize,journal]{IEEEtran}

\usepackage{multirow}
\usepackage{cite}
\usepackage{amssymb}
\usepackage[numbers,sort&compress]{natbib}
\usepackage[caption=false,font=normalsize,labelfont=sf,textfont=sf]{subfig}
\usepackage{threeparttable}
\usepackage{booktabs}
\usepackage{graphicx}
\usepackage{mathtools}
\usepackage{algorithm}
\usepackage{algorithmic}
\usepackage{url} 

\begin{document}

%
\title{Restricted Black-box Adversarial Attack \\ Against DeepFake Face Swapping}
%
%
%

\author{Junhao~Dong,
        Yuan~Wang,
        Jianhuang~Lai,~\IEEEmembership{Senior~Member,~IEEE}
        and~Xiaohua~Xie,~\IEEEmembership{Member,~IEEE}
\thanks{The authors are with the School of Computer Science and Engineering, Sun Yat-sen
	University, Guangzhou 510006, China, also with the Key Laboratory of
	Machine Intelligence and Advanced Computing, Ministry of Education,
	Guangzhou 510006, China, and also with the Guangdong Key Laboratory
	of Information Security Technology, Guangzhou 510006, China
	(e-mail: dongjh8@mail2.sysu.edu.cn; wangy975@mail2.sysu.edu.cn; stsljh@mail.sysu.edu.cn; xiexiaoh6@
	mail.sysu.edu.cn).
	\textit{(Corresponding author: Xiaohua Xie.)}}
}

%
%

\markboth{IEEE TRANSACTIONS ON Information Forensics and Security}
{Junhao Dong \MakeLowercase{\textit{et al.}}: Restricted Black-box Adversarial Attack Against DeepFake Face Swapping}
%



\maketitle

\begin{abstract}
DeepFake face swapping presents a significant threat to online security and social media, which can replace the source face in an arbitrary photo/video with the target face of an entirely different person. In order to prevent this fraud, some researchers have begun to study the adversarial methods against DeepFake or face manipulation. However, existing works focus on the white-box setting or the black-box setting driven by abundant queries, which severely limits the practical application of these methods. To tackle this problem, we introduce a practical adversarial attack that does not require any queries to the facial image forgery model. Our method is built on a substitute model persuing for face reconstruction and then transfers adversarial examples from the substitute model directly to inaccessible black-box DeepFake models. Specially, we propose the Transferable Cycle Adversary Generative Adversarial Network (TCA-GAN) to construct the adversarial perturbation for disrupting unknown DeepFake systems. We also present a novel post-regularization module for enhancing the transferability of generated adversarial examples. To comprehensively measure the effectiveness of our approaches, we construct a challenging benchmark of DeepFake adversarial attacks for future development. Extensive experiments impressively show that the proposed adversarial attack method makes the visual quality of DeepFake face images plummet so that they are easier to be detected by humans and algorithms. Moreover, we demonstrate that the proposed algorithm can be generalized to offer face image protection against various face translation methods.
\end{abstract}


%
\IEEEpeerreviewmaketitle

\section{Introduction}
%
%
%
%
\IEEEPARstart{D}{eepFake}, a portmanteau of ``deep learning'' and ``fake'', has drawn broad attention in recent years. This burgeoning technique can replace the \textit{source} face in an existing image or a video with the \textit{target} face of an arbitrary identity. However, the abuse of this intriguing technology may result in an underlying hazard to social media and online security. For instance, it can be maliciously used to blackmail individuals or to bypass the authentication mechanism \cite{korshunov2018deepfakes}.

Although these subtle DeepFake artifacts confuse human vision, several forgery detection methods can precisely distinguish these manipulated images \cite{rossler2019faceforensics++, afchar2018mesonet, nguyen2019capsule}. However, there exists a time delay between the publishing of forged images and their corresponding detection results, which may damage personal reputation by impersonating the victim on social media platforms. Another way to defend against this malicious face manipulation is to disrupt the generation stage of DeepFake face swapping, \textit{i.e}., to distort generated face swapping images. This straightforward method eliminates the complicated forgery detection, which can tackle these DeepFake issues from the root thoroughly.

\begin{figure}
	\centering
	\includegraphics[width=0.99\linewidth]{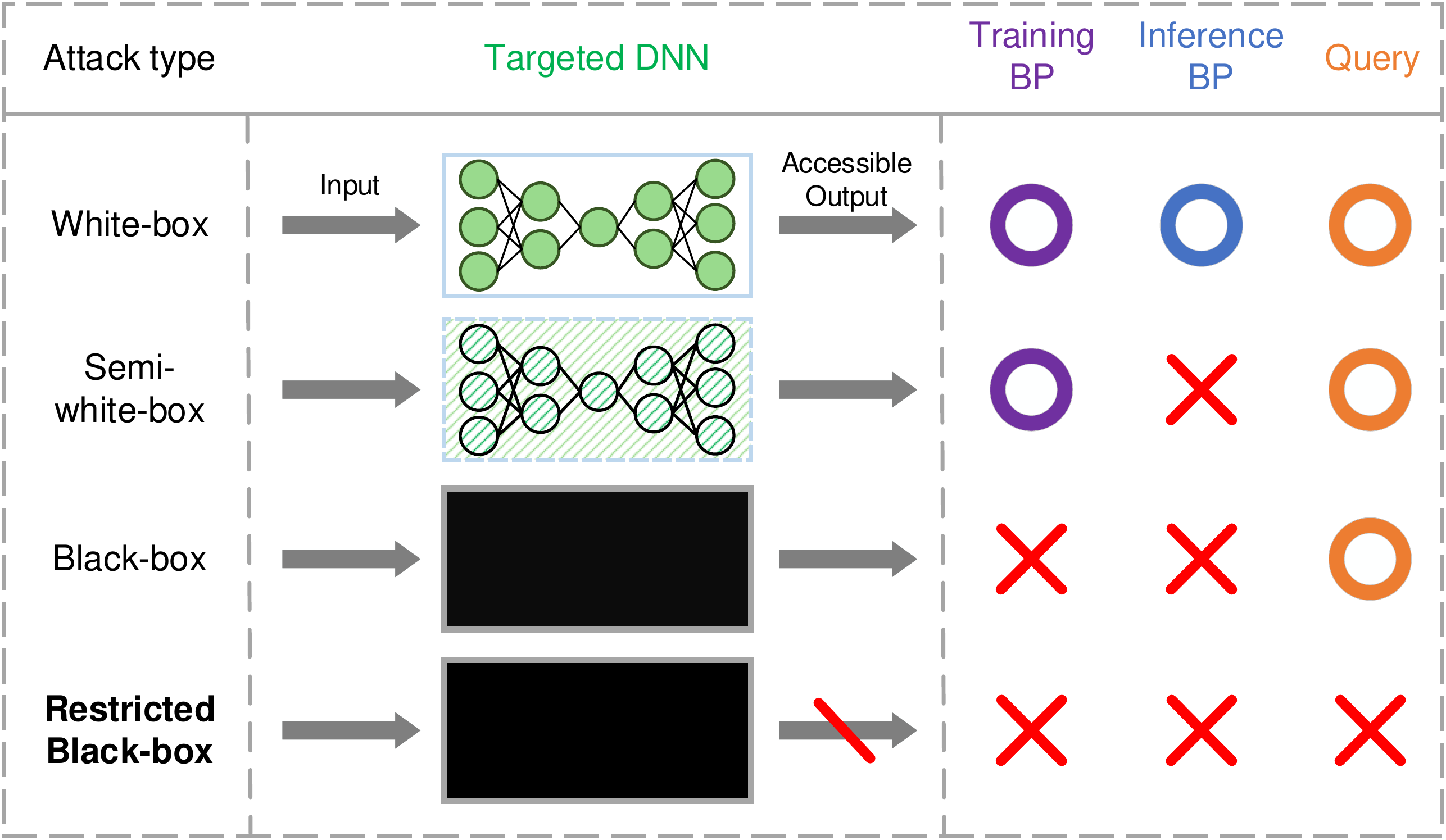}
	\caption{Illustration of the taxonomy of adversarial attacks against the target Deep Neural Network (DNN). ``Training and inference Backward Propagation (BP)'' are corresponding to the accessibility to backward gradients of the target model, which are related to training and inference of our adversary generative model (threat model). ``Query'' indicates the availability of inference outputs of the target model. }
	\label{fig:1}
\end{figure}

Deep Neural Networks (DNNs) have made spectacular progress in various computer vision tasks \cite{He_2016_CVPR,ronneberger2015u,NIPS2014_5423}. Nonetheless, DNNs are especially vulnerable to some tailored examples synthesized by attaching imperceptible perturbations to original images. These visually unconscious examples with the added perturbation are also regarded as adversarial examples \cite{journals/corr/SzegedyZSBEGF13}. Adversarial examples can drastically mislead the inference of DNNs to output wrong (or even specific) results, which poses a significant challenge to current deep learning applications \cite{9325048,9430730,kos2018adversarial}. A plausible reason for the emergence of the adversarial example is the linear behavior in high dimensional feature spaces of DNNs \cite{DBLP:journals/corr/GoodfellowSS14}. Thus, researching on adversarial examples helps explain the learned feature of DNNs and shows a path to attack the improper usage of DNN mechanisms like DeepFake. Significantly, the attack scenarios are various according to the accessible information of the target model and its output results, as shown in Fig. \ref{fig:1}. The white-box adversarial attack has the complete information of the target model. Nevertheless, the semi-whitebox setting \cite{xiao2018generating} focuses on training a DNN to generate adversarial examples, which do not need the backward gradients at the inference stage. However, these two types of attacks are impractical in defending against unknown tampering. The black-box adversarial attack can only gain the DNN output. In contrast, the restricted black-box adversarial attack \cite{chen2017zoo} is built on an entirely black box with the inaccessible output of the target DNN model, which means that we can not conduct even a single query to the black-box model. Therefore, this attack is more applicable to real-world scenarios. To achieve general applications, we consider the restricted black-box adversarial attack, which can defend the immutability of our photos from DeepFake face swapping or other face editing systems.

\begin{figure}
	\centering
	\includegraphics[width=0.99\linewidth]{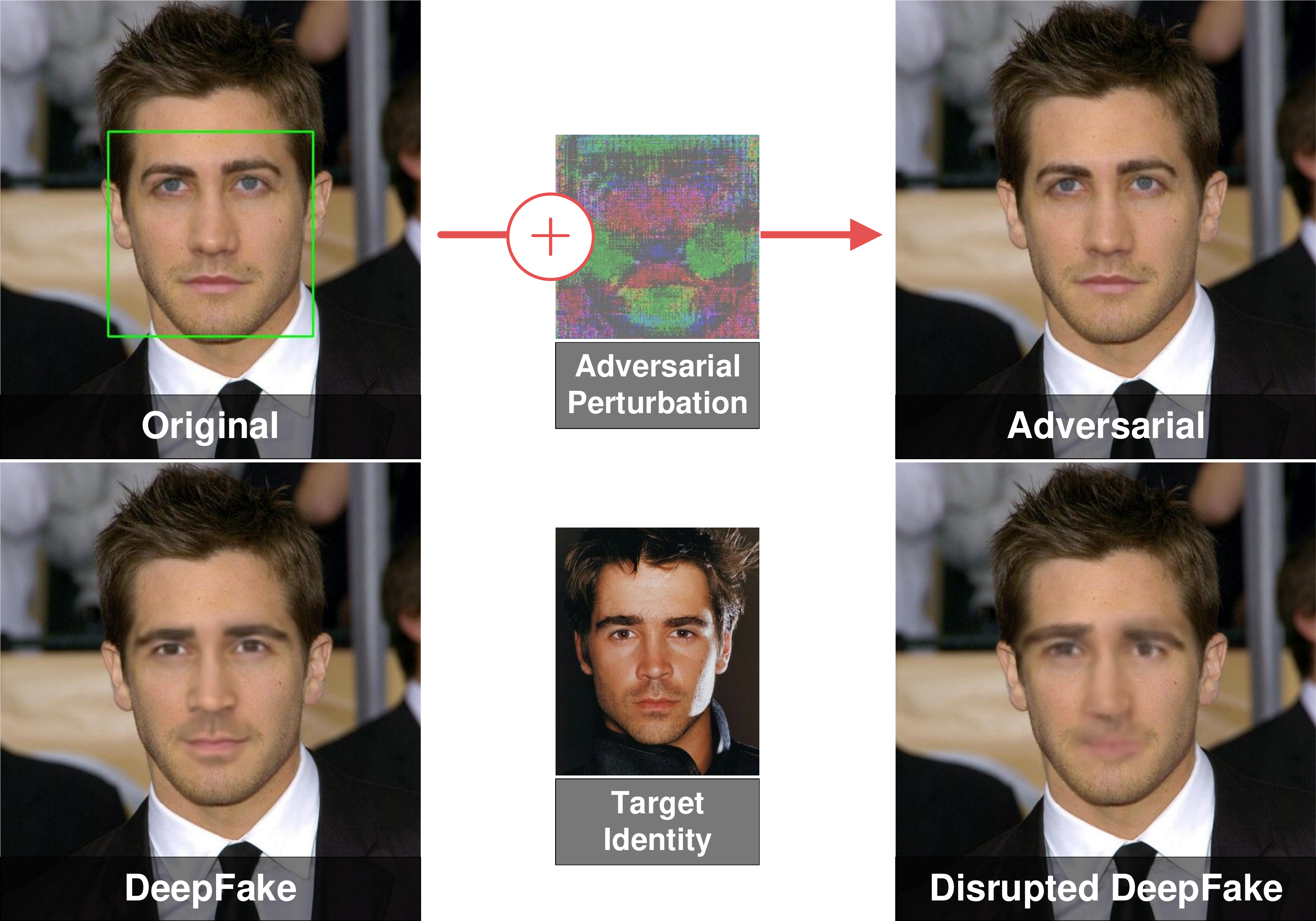}
	\caption{Comparison of DeepFake face swapping between the legitimate and the adversarial example. DeepFake aims to replace the source face in the original image with the fake face of the target identity. The adversarial example can be easily obtained by adding a visually undetectable adversarial perturbation (magnified for visibility) to the original image. Note that this tailored adversarial example is visually indistinguishable from the original example in the human vision, but can result in a strong disruption to the output face-swapped image.}
	\label{fig:2}
\end{figure}

In this paper, we consider constructing a substitute model to simulate the process of face reconstruction, which is achieved by a deep autoencoder network. The main reason is that we consider DeepFake face swapping, face style manipulation, \textit{etc.} as approximations of face reconstruction. Then, we conduct adversarial attack towards the accessible substitute model and then transfer adversarial examples to other unknown face manipulation models. Specifically, we innovatively build the Transferable Cycle Adversary Generative Adversarial Network (TCA-GAN) to generate adversarial perturbations against unknown DeepFake face swapping systems. The training procedure is mainly based on adding and removing the adversarial perturbation to simulate both the adversarial attack and defense. In the application stage, TCA-GAN can generate the adversarial perturbation against the unknown face swapping model from a single face image. The generalizable adversarial example can be obtained by adding the adversarial perturbation to the input legitimate face image, which can disrupt the DeepFake face swapping or even other face manipulation models. For better generalization results, we also apply a novel post-regularization on the generated adversarial example to enhance its generalizability. We show an illustrated example of DeepFake face swapping with respect to the original image and the adversarial example in Fig. \ref{fig:2}.

To comprehensively validate our proposed method, we build a benchmark of DeepFake adversarial attacks for future research. Moreover, We evaluate the face swapping results on both referenced and non-referenced image quality assessments. Extensive experiments impressively show that the proposed method can degrade the visual quality of DeepFake face images effectively so that they are easier to be identified by human eyes and DeepFake detectors. We also demonstrate that the proposed method can be well generalized to offer protection against several face translation methods with nearly no cost.

Overall, our main contributions can be summarized as follow: 
\begin{itemize}
	\item To the best of our knowledge, we are the first to successfully construct a practical restricted black-box attack against DeepFake face swapping. Specifically, we build the face reconstruction autoencoder as the substitute model for producing adversarial examples that can be directly transferred to unknown face manipulation models. 
	\item We design TCA-GAN to generate powerful adversarial examples against DeepFake systems. Moreover, the post-regularization module is proposed for better transferability. Extensive experiments show that our method can be generalized to various face attribute editing models with nearly no cost. 
	\item We construct strong benchmarks on disrupting DeepFake systems when simulating a real-world scenario of arbitrary face swapping. In addition to disrupting the visual quality of generated fake images, we demonstrate that the disruption can further enhance the performance of several image-level DeepFake detection methods.
\end{itemize}

\section{Related works}
\subsection{Automatical Face Swapping}
Face swapping consists of replacing the face of a person with another face of a different identity. This tricky technique was first conducted in \cite{blanz2004exchanging}, which applies a 3D morphable model fitting algorithm on both exchanging faces. Bitouk \textit{et al.} \cite{bitouk2008face} presented a fully automatic face swapping by replacing the source face with the candidate face image from a face gallery. Nevertheless, the target identity of this face swapping is uncontrollable and time-consuming. Korshunova \textit{et al.} \cite{8237659} regarded the face swapping problem as style transfer to accomplish designated face swapping, which is conducted by a deep convolutional neural network with efficient preprocessing and postprocessing. Especially, Bao \textit{et al.} \cite{bao2018towards} disentangled the identities and attributes of face images and proposed an identity preserving GAN-based network for open-set face synthesizing. Modern deep learning-based face swapping methods eliminate several fussy face editing steps and synthesize photo-realistic face images \cite{li2019faceshifter,nirkin2019fsgan,chen2020simswap}. The most popular technique of these is DeepFake \cite{korshunov2018deepfakes}, which induces a plethora of fake public scandals. However, numerous researches have been explored on DeepFake detection. We consider a straightforward approach to disturbing DeepFake face swapping.  

\begin{figure*}[h!]
	\centering
	\includegraphics[width=0.99\linewidth]{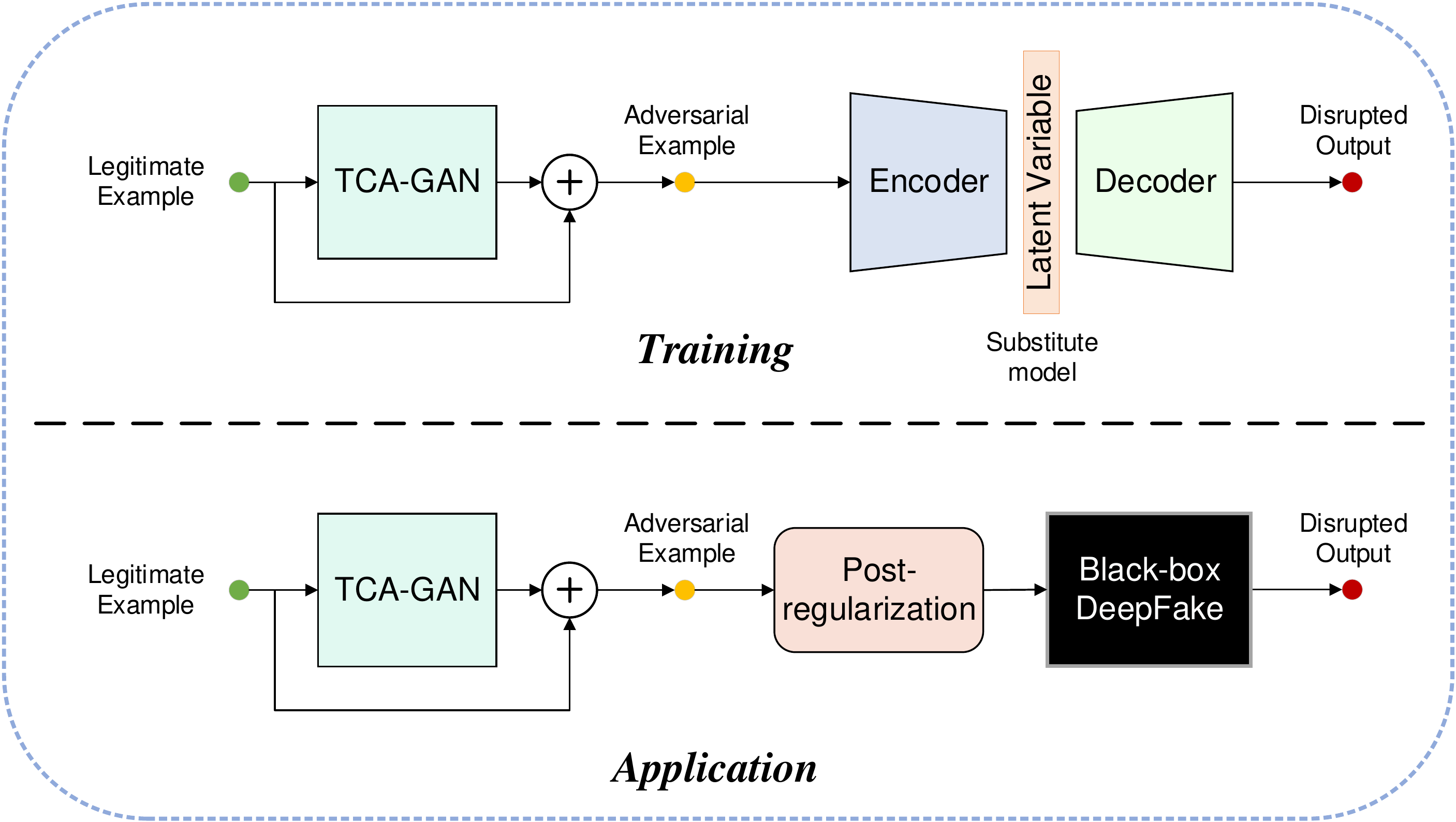}
	\caption{The flow chart of our proposed method. Due to the accessibility to the substitute model in the training stage, we train TCA-GAN to produce the adversarial perturbation against the substitute model according to the input face image. In the application stage, we append a post-processing module to regularize the generated adversarial example for better transferability. The application stage simulates a real-world scenario of attacking DeepFake, while we can not obtain the face-swapped output or any details of DeepFake.}
	\label{fig:3}
\end{figure*}

\subsection{Adversarial attacks against classifiers}
As demonstrated by \cite{journals/corr/SzegedyZSBEGF13}, the reason for adversarial examples is the discontinuities of DNNs. Alternatively, Goodfellow \textit{et al.} \cite{DBLP:journals/corr/GoodfellowSS14} argued that adversarial examples are results from the linearity in high dimensional space in DNNs and presented the Fast Gradient Sign Method (FGSM). The iterative strategy is also exerted into FGSM to build more powerful adversarial attacks \cite{kurakin2016adversarial,madry2017towards}. Moosavi-Dezfooli \textit{et al.} \cite{moosavi2017universal} extend adversarial attack to an universal (image-agnostic) scenario, \textit{i.e.}, making wrong prediction for each input image with high confidence. Especially, Su \textit{et al.} \cite{8601309} considered an extremely confined scenario and proposed a one-pixel adversarial perturbation generative method based on differential evolution. In addition, Croce \textit{et al.} \cite{croce2020minimally} introduced a Fast Adaptive Boundary (FAB) attack to find minimal adversarial perturbation in distinct norm constraints. Particularly, Xiao \textit{et al.} \cite{xiao2018generating} proposed AdvGAN to generate more effective adversarial examples against the classification with GANs. Our method also focus on the same GAN-based adversary generation. However, we extend it to attacking generative models in the restricted black-box setting via constructing a robust cycle-consistent GAN. Zhang \textit{et al.} \cite{9186644} studied the optimization on the manifold of the classification boundary and introduced boundary projection to obtain low adversarial perturabtion. Zhong \textit{et al.} \cite{9252132} proposed a novel surrogate model to enhance the transferability of adversarial attacks against face recognition models. Likewise, our goal is to enhance the transferability of adversarial examples, while we mainly focus on adversarial examples against DeepFake face swapping to protect  in the restricted black-box scenario. While our focus is not limited to adversarial attacks against malicious face swapping to protect the security of personal photos, it yields supportive generalization performance on attacking several face manipulation models as well.

\subsection{Adversarial attacks against generative models}
\vspace{-1pt}
Although adversarial attacks towards classification models have achieved satisfactory performance, few studies have focused on adversarial attacks against generative models. Tabacof \textit{et al.} \cite{tabacof2016adversarial} first investigated adversarial attacks for autoencoders with latent representation. They indicated that generative models do not have a definite decision boundary as classification models and thus are more difficult to disrupt. Kos \textit{et al.} \cite{kos2018adversarial} attached an auxiliary classifier to the encoder of the target generative models to obtain classification-based adversarial attacks. Recently, applying adversarial examples to disturb image-translation models receives heavy investigations \cite{ruiz2020disrupting, ruiz2020protecting, fang2020adversarial}. Particularly, Dong \textit{et al.} \cite{Dong_deepfake} demonstrate impressive results on disrupting DeepFake face swapping in white-box circumstances. In contrast, we explore the restricted black-box scenario. Sun \textit{et al.} \cite{sun2020landmark} described a white-box adversarial attack to disturb the facial landmark extraction, thus can induce the misalignment on DeepFake face swapped images. Their method mainly focuses on the preprocessing of DeepFake face swapping, while we concentrate on a 	universal way to disturb the process of face swapping. Ruiz \textit{et al.} \cite{ruiz2020protecting} first construct a multi-query based adversarial attack against image-translation models in the black-box scenario. Nevertheless, their black-box scenario is conducted through numerous queries, which is impractical to defend the malicious face swapping in real-world conditions. In comparison, our method considers a more practical restricted black-box setting, with no need for queries to the target face swapping model.

\section{Method}
\subsection{Preliminaries}
\subsubsection{Brief review of DeepFake}
Original DeepFake consists of one shared encoder and two separate decoders to two face swapping identities respectively. The shared encoder converts the input face image into a latent representation. Then the decoder transforms the latent representation to a face image, whose identity is according to the selected decoder. In the training stage, DeepFake conducts face reconstruction with the decoder that is consistent with the input identity. For face swapping, DeepFake decodes the latent variable with the other decoder to obtain face swapping images and vice versa. Note that these input face images have already been trained for robust face swapping. Hence, it is also challenging to disrupt DeepFake face swapping. In this paper, we mainly focus on attacking this original type of DeepFake face swapping mechanism.

\subsubsection{Notation}
In this paper, we intend to build generalizable adversarial examples against DeepFake face swapping in the black-box scenario. We denote the adversarial example as $x^{adv}= x + r$, where $x$ is the legitimate input face image, and $r$ is the appended adversarial perturbation. Let $DF: X \rightarrow Y$ stand for the inaccessible DeepFake face swapping model, which maps the input face image $x\in X$ to a face-swapped output $DF\left( x\right) \in Y$. Note that $X$ and $Y$ are corresponding to the domain of \textit{source} images and face-swapped \textit{target} images, respectively. The substitute model $S\left( \cdot \right) = S_{d}\left( S_{e}\left( \cdot \right) \right)$ is composed of a downsampling encoder $S_{e}\left( \cdot \right) $ and an upsampling decoder $S_{d}\left( \cdot \right) $.

\begin{figure*}[h!]
	\centering
	\includegraphics[width=0.99\linewidth]{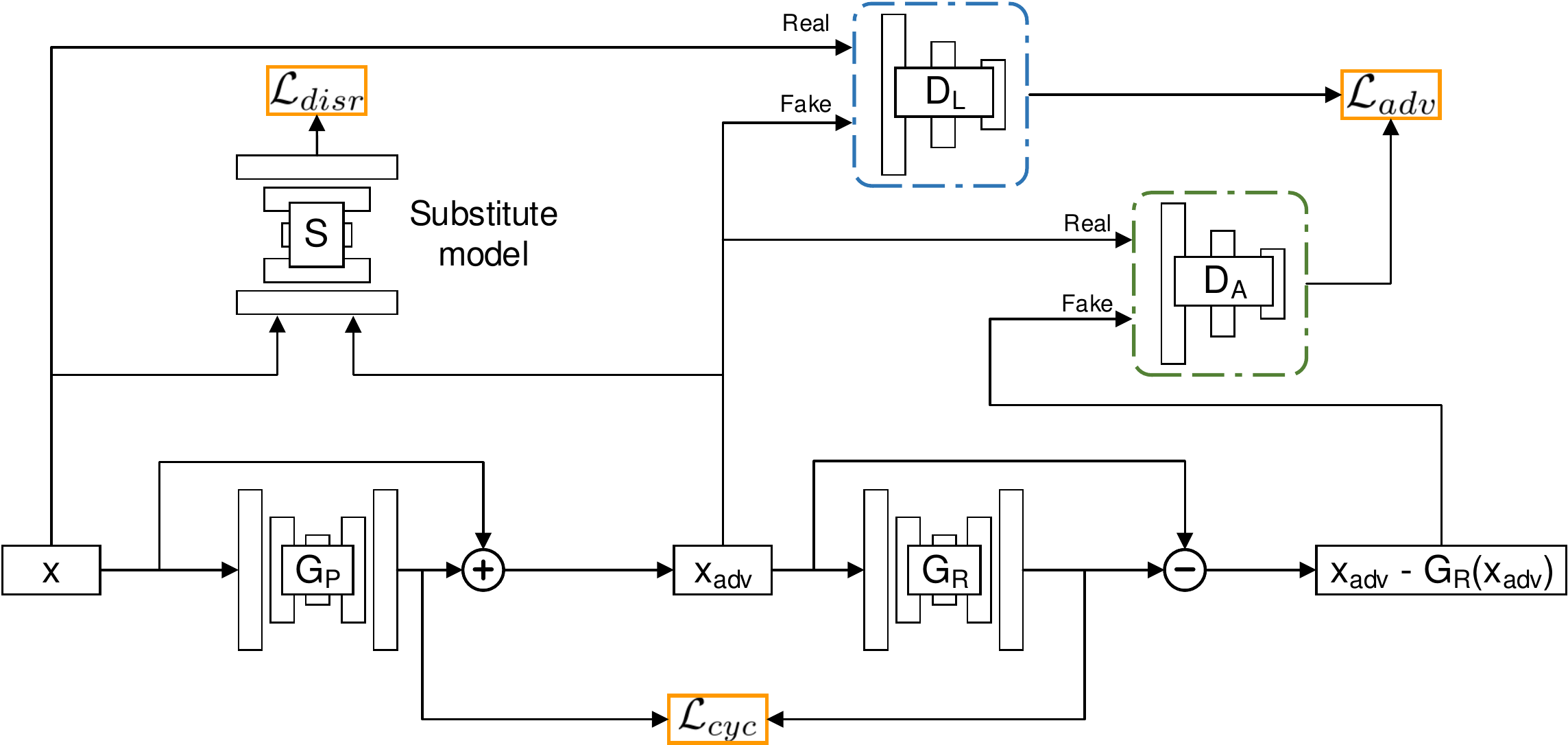}
	\caption{Illustration of the training mechanism of TCA-GAN, including two generators and two domain discriminators. These two generators are utilized to produce and remove the adversarial perturbation. The corresponding discriminators differentiate both legitimate examples and adversarial examples. }
	\label{fig:4}
\end{figure*}

Our goal is to find a transferable adversarial perturbation to disrupt the target DeepFake model simultaneously with no accessing to the DeepFake model. More pertinently, we expect to enhance the difference between face-swapped images generated by the legitimate example and the adversarial example respectively. We first formulate the pursued adversarial perturbation as follow:
\begin{equation}
\begin{aligned}
&\max\limits_{r} \quad   F \left [ DF\left ( x \right ) , DF\left ( x+r \right ) \right]  \\
& \begin{array}{r@{\quad}r@{}l@{\quad}l}
s.t.&\left \| r \right \|_{\infty} &\leq \epsilon \\
\label{eq:1}
\end{array}
\end{aligned}
\end{equation}
where $F$ is the metric to measure the distance between the face-swapped images, $\epsilon$ is the infinity norm bound to restrict the adversarial perturbation.

\subsection{TCA-GAN against the substitute model}


As aforementioned, query-based black-box adversarial attack is limited by abundant query times to the black-box model during inference stage, which need to have the access to the target black-box model. However, we can not conduct multiple query to black-box models or even access the output of them in some real-world scenarios. Consequently, we consider a restricted black-box adversarial attack with no access to the target DeepFake model. According to the transferability of adversarial examples \cite{DBLP:journals/corr/GoodfellowSS14}, we distinctly build a substitute model for generalizable adversary generation. More specifically, we obtain adversarial examples from the substitute model and expect to transfer them to the unknown DeepFake face swapping model. Here, the adversarial example is composed of a legitimate image and the adversarial perturbation that is generated by TCA-GAN with the prior of legitimate image. Then we apply a post-regularization on the adversarial example to enhance the transferability. The flow chart of our adversarial attack is shown in Fig. \ref{fig:3}. On the whole, we acquire the adversarial example against the substitute model through TCA-GAN. Afterward, we can transfer it to the target DeepFake face swapping model.

Motivated by \cite{papernot2017practical}, we extend the substitute model to the generative model to simulate the black-box DeepFake model, which is much more harder then the classification model that has an absolute classification boundary. Considering that DeepFake and other face manipulation models can be virtually regarded as rebuilding the objective face, we employ a DNN autoencoder accomplishing face reconstruction to build the substitute model. Then the adversarial example disrupting the substitute model can also be transferred to perturb face manipulation models such as DeepFake. The optimization procedure can be formalized as bellow:

\begin{equation}
\begin{aligned}
\mathcal{L}_{recons} = &\left \| S\left( x \right)  - x \right \|_{1} + 	\left \| S\left( \hat{x} \right)  - \hat{x} \right \|_{1} 
\label{eq:2}
\end{aligned}
\end{equation}
where $\hat{x}$ is a slightly wrapped version of the input face image $x$. This wrapping operation comprises randomly rotation, scaling and shifting. The reason we take the wrapped face image into consideration is that the substitute model should possess a certain extent of robustness towards some subtle perturbations. Therefore, the adversarial example that produced against the substitute model can present a better transferring performance on the target model.

After getting the trained substitute model, we then construct TCA-GAN to generate the adversarial perturbation to fool this model. TCA-GAN is a cyclic structure, which mainly consists of two generative modules and two corresponding domain discriminators. These two cycle-consistent adversary generators separately produce and remove the adversarial perturbation. Note that the generated adversarial perturbation can be further added to the original image to get the adversarial example. Respectively, two domain discriminators and are utilized to distinguish the legitimate example $x$ and the adversarial example $x^{adv}$. The mechanism of TCA-GAN is presented in Fig. \ref{fig:4}. 

The forward adversary generation part is to produce an adversarial perturbation to combine with the input face image, disrupting the reconstruction of the substitute model. The generated adversarial examples not only disrupt the ultimate reconstruction of the substitute model but also perturb the latent variable in the substitute model. This intuition is based on the observation that adversarial perturbations against DNNs are transferable \cite{inkawhich2019feature}. The similarity between the components of DNNs incur similar feature representations between different DNNs. Hence, the obtained adversarial example is prone to be transferred to attack the unaccessible DeepFake model. The loss function of disturbing the reconstruction can be written as below:

\begin{equation}	
\footnotesize
\begin{aligned}
\mathcal{L}_{disr} = &\exp \left\{ - \left\| S_{e}\left( x \right)  -  S_{e}\left( G_{P}\left( x \right) + x \right)  \right\|_{1}  \right\} \\
+ &\exp \left\{ - \left\| S\left( x \right)  - S\left( G_{P}\left( x \right) + x \right)  \right\|_{1}  \right\} 
\label{eq:3}
\end{aligned}
\end{equation}
where $G_{P}\left( \cdot \right)$ is the generator to produce the transferable adversarial perturbation with the prior of an legitimate face image. Eq. \ref{eq:3} can also be considered as maximizing the reconstruction results between the adversarial example and the legitimate example. Oppositely, the reversed adversary removal part is to get rid of the appended adversarial perturbation from an adversarial example, \textit{i.e.}, to extract the adversarial perturbation from the input example. Here, we establish a cycle-consistent structure with the two-way adding or removing the adversarial perturbation. More formally, we opt to optimize:

\begin{equation}
\mathcal{L}_{cyc} = \left\|  G_{R}\left( x + G_{P}\left( x \right) \right) - G_{P}\left( x \right)  \right\|_{1} 
\label{eq:4}
\end{equation}
where $G_{R}\left( \cdot \right)$ is the generator to remove the adversarial perturbation from an input adversarial example. The intuition behind this mechanism is that the adversarial perturbation generated to append and remove should be as closer as possible. Hence, it can bring richer supervision on generating puissant adversarial perturbations. This cycle-consistent structure can further enhance the generalizability performance of the adversarial examples.

To balance generators relatively, we construct two domain discriminators to distinguish adversarial examples and legitimate examples. The adversarial loss function of our GAN structure can be formalized as follow:

\begin{equation}
\begin{aligned}
\mathcal{L}_{adv} &= D_{L}(x_{adv}) - D_{L}(x) \\
&+ D_{A}(x_{adv} - G_R(x_{adv})) - D_{A}(x_{adv})
\label{eq:5}
\end{aligned}
\end{equation}
where $D_{L}(\cdot)$ denotes the discriminator of the domain on legitimate examples and $D_{A}(\cdot)$ represents the discriminator of the domain on adversarial examples. For brevity, $x_{adv} = x + G_P(x)$ is the adversarial example created by the adversary generator. Consequently, the adversarial reaction will further enhance the performance of generators until Nash equilibrium. Overall, the total objective of TCA-GAN can be formalized as below:

\begin{equation}
\mathcal{L} = \mathcal{L}_{adv} + \lambda_{cyc}  \mathcal{L}_{cyc} + \lambda_{disr}  \mathcal{L}_{disr}
\label{eq:6}
\end{equation}
where $\lambda_{cyc}$ and $\lambda_{disr}$ manage the relative significance of the objective function. We conduct the GAN mechanism via solving the min-max game, in which we minimize the objective function when training the generator and maximize it when training the discriminator.

\subsection{Post-regularization}
In order to further enhance the transferability of our adversarial examples, we append a post-regularization to weaken their specificity on the substitute model. Inspired by the observation from \cite{huang2019enhancing, Dong_deepfake} that the more specific adversarial examples targeting against the deep learning model, the less generalizable they become. We then slightly shift the attention of generated adversarial examples away from the substitute model, which means that we aim at obtaining a second-best adversarial example towards the substitute model for better generalization.

One effective way to regularize the off-the-shelf adversarial example is to make a distillation. As aforementioned, we disrupt the substitute model by increasing the discrepancy of latent variables between the legitimate example and the adversarial example. We first initialize the new adversarial example by appending random noise to e the non-smooth vicinity of the original example. Then we guide this new adversarial example to approximate the adversarial example generated by TCA-GAN. The corresponding optimization can be written as follow:

\begin{equation}
\begin{aligned}
&\max\limits_{x_{radv}} \quad [S_{e}(x_{radv}) - S_{e}(S(x))] \circ [S_{e}(x_{adv}) - S_{e}(S(x))]\\
& \begin{array}{r@{\quad}r@{}l@{\quad}l}
s.t.&\left\| x_{radv} - x \right\|_{\infty} < \epsilon
\label{eq:7}
\end{array} 
\end{aligned} 
\end{equation}
where $x_{radv}$ represents the regularized adversarial example and $\circ$ is the hadamard product. Note that the consequent term in Eq. \ref{eq:7} is served as a weight on increasing latent discrepancy during the optimization. As aforementioned, the post-regularization can be regarded as a distillation from the adversarial example generated from TCA-GAN. Due to the consistency before and after the reconstruction of the autoencoder, we adopt to maximize the feature-level distance between regularized adversarial examples and reconstructed examples. The reason why we choose to involve reconstructed examples but not original examples is that reconstructed examples can be viewed as an augmentation of original examples. Thus we can obtain more generalizable adversarial examples through attacking against both original example and their augmented versions simultaneously. This optimization can also induce the regularized adversarial example to be more robust to small transformation. Furthermore, the regularized adversarial example is optimized to keep the same high latent discrepancy. To simplify this optimization procedure, we adopt an iterative gradient ascent strategy on the regularized adversarial example. Afterward, we project the regularized adversarial example to a restriction to limit its magnitude. The overall procedure is shown in Algorithm \ref{alg:1}, where $Clip_{\epsilon}$ and $Clip_{image\_range}$ represents the clipping in the $\epsilon$-ball and the image range, respectively. Note that the randomization in the initial stage of adversarial examples can further prevent the label leaking and gradient masking problem.

\begin{algorithm}[tb] 
	\caption{Post-regularization} 
	\label{alg:1} 
	\begin{algorithmic}[1] 
		\REQUIRE ~~\\ 
		Adversary generator of TCA-GAN $G_P$;\\
		Original face image $x$;\\
		Substitute model $S$;\\
		Adversarial perturbation bound $\epsilon$;\\
		Iterative number $N$;\\
		Iterative step size $\alpha$;\\
		\ENSURE ~~\\ 
		Regularized adversarial perturbation $x_{radv}$;
		\STATE $x_{adv} = x + G_{P}(x)$;
		\STATE Randomly initialize $x_{radv}$ from the neighborhood of $x_{adv}$;
		\STATE $x_{rec} = S(x)$
		\STATE $W = S_{e}(x_{adv}) - S_{e}(x_{rec})$
		\FOR {$k=1,\ldots,N$}
		\STATE $L = [(S_{e}(x_{radv}) - S_{e}(x_{rec})) \circ W] \ \big / \ \left\| x \right\|_{F} $;
		\STATE $r = Clip_{\epsilon}(x_{radv} - x + \alpha sign(\nabla_{x_{radv}}L))$;
		\STATE $x_{radv} = Clip_{image\_range}(x + r)$
		\ENDFOR  \\
		\RETURN $x_{radv}$. 
	\end{algorithmic}
\end{algorithm}

\section{Experiments}

In this section, we first introduce the proposed benchmarks on disrupting DeepFake face swapping and the measurement of our experimental results. Next, we make a comparison with other transferrable adversarial attack methods and conduct an ablation study on the component modules. To further explore the effectiveness of our method, we verify that the perturbed face-swapped images can facilitate DeepFake detection. Comprehensively, we also apply our method on several face manipulation models to validate the generalization performance.

\subsection{Experimental Setup}
\subsubsection{Dataset}
Following \cite{Dong_deepfake}, we construct a larger DeepFake face swapping dataset of 6274 images, including 40 man identities and 38 woman identities. To support future development of DeepFake adversarial attacks, we also train the corresponding DeepFake face swapping models on the proposed dataset. Note that in order to simulate the real-world application scenario, the database to train both the substitute model and TCA-GAN is disjoint with the database on disrupting DeepFake face swapping. Moreover, CelebA \cite{liu2015deep} dataset is utilized for verifying the transfer-based adversarial attack on other face manipulation models.

\subsubsection{Metrics}
To comprehensively evaluate our proposed method, we conduct both referenced and non-referenced image quality assessments on the face-swapped images. Structure SIMilarity (SSIM) \cite{wang2004image} index and Feature SIMilarity (FSIM) \cite{zhang2011fsim} are measured on both pairs of input face images and face swapped images in terms of legitimate examples and adversarial examples. Besides, Blind/Referenceless
Image Spatial QUality Evaluator (BRISQUE) \cite{mittal2012no} is also utilized to assess the image quality. A smaller BRISQUE score represents better visual quality.

\begin{table}[t!]
	\centering
	\renewcommand{\arraystretch}{1.4}
	\renewcommand{\tabcolsep}{2.1mm}
	\resizebox{0.95\linewidth}{!}{
		\begin{threeparttable}
			\caption{ Reference image quality assessment on disrupting DeepFake face swapping in the black-box setting. The best performance is marked in \textbf{bold}.}
			\begin{tabular}{c||c|c|c|c}
				\toprule
				\multirow{2}{*}{Method}&
				\multicolumn{2}{c}{Source}&\multicolumn{2}{c}{\footnotesize Face swapping} \cr
				\cmidrule(lr){2-3} \cmidrule(lr){4-5}
				& SSIM$\uparrow$ & FSIM$\uparrow$   &SSIM$\downarrow$ & FSIM$\downarrow$  \cr
				\midrule		
				FGSM \cite{DBLP:journals/corr/GoodfellowSS14} &\textbf{0.953}
				&0.975&0.811&0.917\cr
				PGD \cite{madry2017towards}&0.947&0.969&0.788&0.899\cr
				
				DIM \cite{xie2019improving}&0.952&0.973&0.756&0.881\cr
				TIM \cite{dong2019evading}&0.95&\textbf{0.978}&0.76&0.883\cr
				ILA \cite{huang2019enhancing}&0.943&0.967&0.752&0.879\cr
				APGD \cite{croce2020reliable}&0.939&0.965&0.75&0.874\cr
				
				\cmidrule{1-5}
				\textbf{ours}&0.951&0.974&\textbf{0.731}&\textbf{0.873}\cr
				
				\bottomrule
			\end{tabular}
			\label{tab:1}
		\end{threeparttable}
	}
\end{table}

\subsubsection{Implementation Details}
%
For the purpose of simulating the practical adversarial attack scenario, the generated adversarial examples will be further resized to the original size. Consequently, they will be resized and randomly transformed before entering the target DeepFake model. In this paper, the image value is normalized to $\left[ 0, 1 \right]$. To obtain the visually undetectable adversarial perturbation, we restrict the bound $\epsilon$ to be $0.03$. We run the post-regularization for $10$ iterations with a $0.006$ iterative step size.

\subsection{Experimental Results}
\begin{figure}[t!]
	\centering
	\subfloat[]{
		\begin{minipage}[t]{0.22\linewidth}
			\centering
			\includegraphics[width=0.99\linewidth]{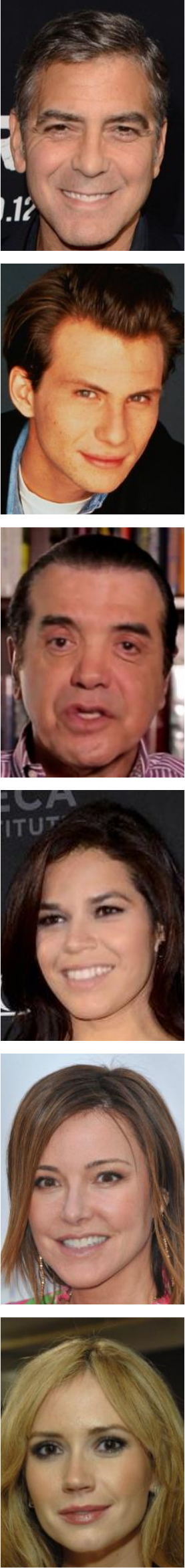}
		\end{minipage}%
	}
	\subfloat[]{
		\begin{minipage}[t]{0.22\linewidth}
			\centering
			\includegraphics[width=0.99\linewidth]{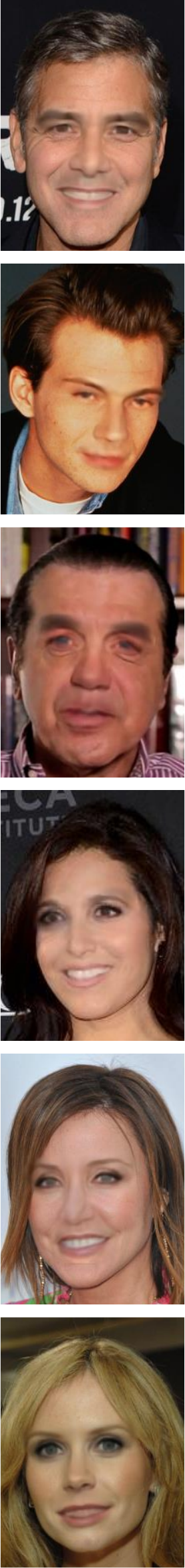}
		\end{minipage}%
	}
	\subfloat[]{
		\begin{minipage}[t]{0.22\linewidth}
			\centering
			\includegraphics[width=0.99\linewidth]{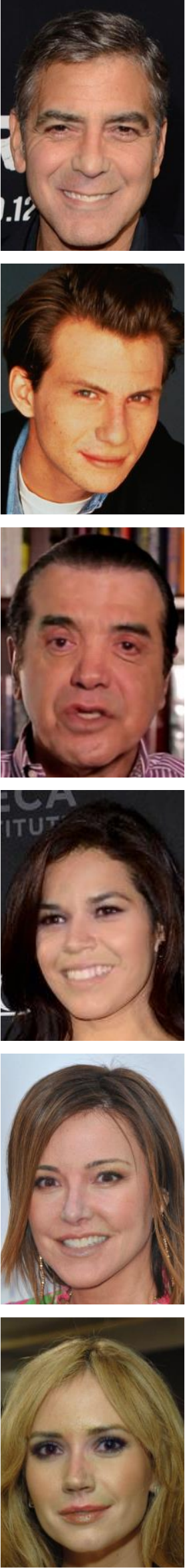}
		\end{minipage}%
	}
	\subfloat[]{
		\begin{minipage}[t]{0.22\linewidth}
			\centering
			\includegraphics[width=0.99\linewidth]{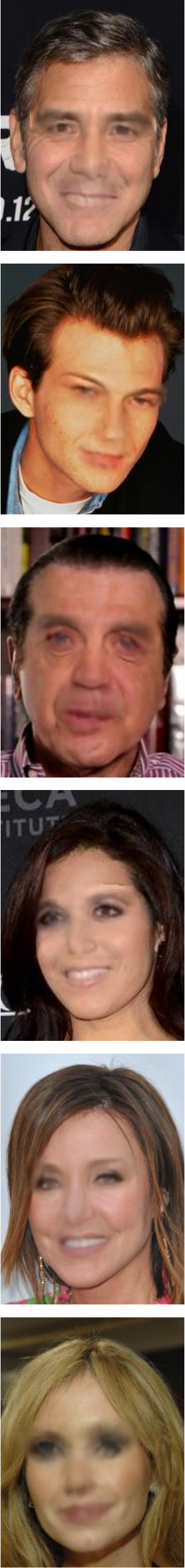}
		\end{minipage}%
	}
	\centering
	\caption{Examples of both original and disrupted face swapping in the restricted black-box scenario. DeepFake model generates original face-swapped images (b) and disrupted face-swapped images(d) with input of legitimate examples (a) and adversarial examples (c), respectively. }
	\label{fig:5}
\end{figure}

To begin with, we present several face swapping examples in Fig. \ref{fig:5} depending on the input of legitimate images and adversarial images. We perform a sequence of experiments to provide empirical evidence of our method on attacking against DeepFake face swapping. Note that all experiments in this section are conducted in the restricted black-box scenario, \textit{i.e.}, the target DeepFake model is inaccessible until the completion of adversary generation. Moreover, the produced adversarial examples are directly fed in the target DeepFake face swapping models. The comparison with other adversarial attack methods is reported in Table \ref{tab:1}.

The disruption against DeepFake face swapping is evaluated through the similarity between face-swapped images that are generated from legitimate examples and adversarial examples, respectively. Due to the infinite norm restriction on the adversarial perturbation, there remains only a tiny gap in the visual distances between legitimate examples and adversarial examples during the comparison. According to the table, we can observe that our method presents an effective perturbation on DeepFake face swapping while preserving the adversarial perturbation visually imperceptible. Moreover, we evaluate the non-referenced image quality assessment BRISQUE as shown in Table \ref{tab:2}. Note that our goal is to obtain inferior face-swapped images that are induced by adversarial examples while keeping the image quality of input adversarial examples high. Both the referenced and non-referenced image quality assessments demonstrate that our methods can effectively give rise to a facial distortion on DeepFake face-swapped images.

\begin{table}[tb!]
	\centering
	\renewcommand{\arraystretch}{1.4}
	\renewcommand{\tabcolsep}{2.1mm}
	\resizebox{0.95\linewidth}{!}{
		\begin{threeparttable}
			\caption{Non-reference image quality assessment BRISQUE on related images. The best performance is marked in \textbf{bold}.}
			\begin{tabular}{c||c|c|c|c}
				\toprule
				\multirow{2}{*}{Method}&
				\multicolumn{2}{c}{Source images}&\multicolumn{2}{c}{Face-swapped images} \cr
				\cmidrule(lr){2-3} \cmidrule(lr){4-5}
				& Original & Adversarial&Original & Adversarial\cr
				\midrule		
				FGSM \cite{DBLP:journals/corr/GoodfellowSS14}&\multirow{6}{*}{23.58}&\textbf{22.93}&\multirow{6}{*}{35.68}&38.33\cr
				PGD \cite{madry2017towards}&&25.28&&40.74\cr
				DIM \cite{xie2019improving}&&24.17&&44.82\cr
				TIM \cite{dong2019evading}&&24.22&&44.17\cr
				ILA \cite{huang2019enhancing}&&27.04&&44.63\cr
				APGD \cite{croce2020reliable}&&24.12&&44.26\cr
				
				\cmidrule(lr){1-1} \cmidrule(lr){3-3} \cmidrule(lr){5-5}
				\textbf{ours}&&23.97&&\textbf{47.13}\cr
				\bottomrule
			\end{tabular}
			
			\label{tab:2}
		\end{threeparttable}
	}
\end{table}

\begin{table}[t!]
	\centering
	\renewcommand{\arraystretch}{1.4}
	\renewcommand{\tabcolsep}{2.1mm}
	\resizebox{0.99\linewidth}{!}{
		\begin{threeparttable}
			\caption{ Ablation study on the cyle-consistency type in TCA-GAN. The best performance is marked in \textbf{bold}.}
			
			\begin{tabular}{c||c|c|c}
				\toprule
				Cyclic type& SSIM$\downarrow$ & FSIM$\downarrow$ & BRISQUE$\uparrow$\cr
				\midrule		
				w/o cycle-consistency &0.752
				&0.878&45.43\cr
				w/ bidirectional cycle-consistency &0.738
				&0.876&46.07\cr
				w/ unidirectional cycle-consistency &\textbf{0.731}
				&\textbf{0.873}&\textbf{47.13}\cr
				
				\bottomrule
			\end{tabular}
			
			\label{tab:3}
		\end{threeparttable}
	}
\end{table}

\begin{figure*}[ht!]
	\centering
	\subfloat[Original face translation]{
		\begin{minipage}[t]{0.48\linewidth}
			\centering
			\includegraphics[width=0.94\linewidth]{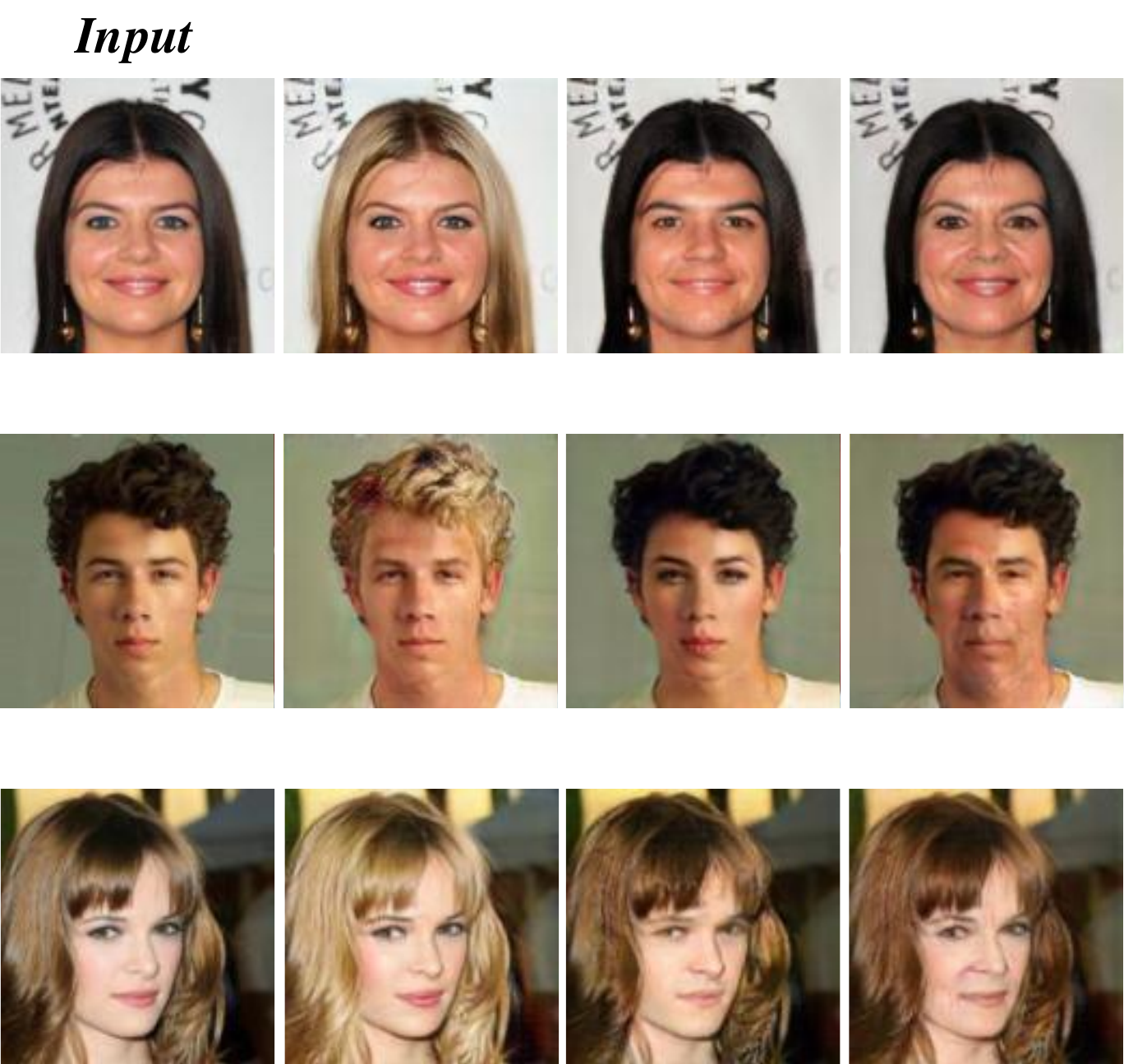}
		\end{minipage}%
	}
	\subfloat[Disrupted face translation]{
		\begin{minipage}[t]{0.48\linewidth}
			\centering
			\includegraphics[width=0.94\linewidth]{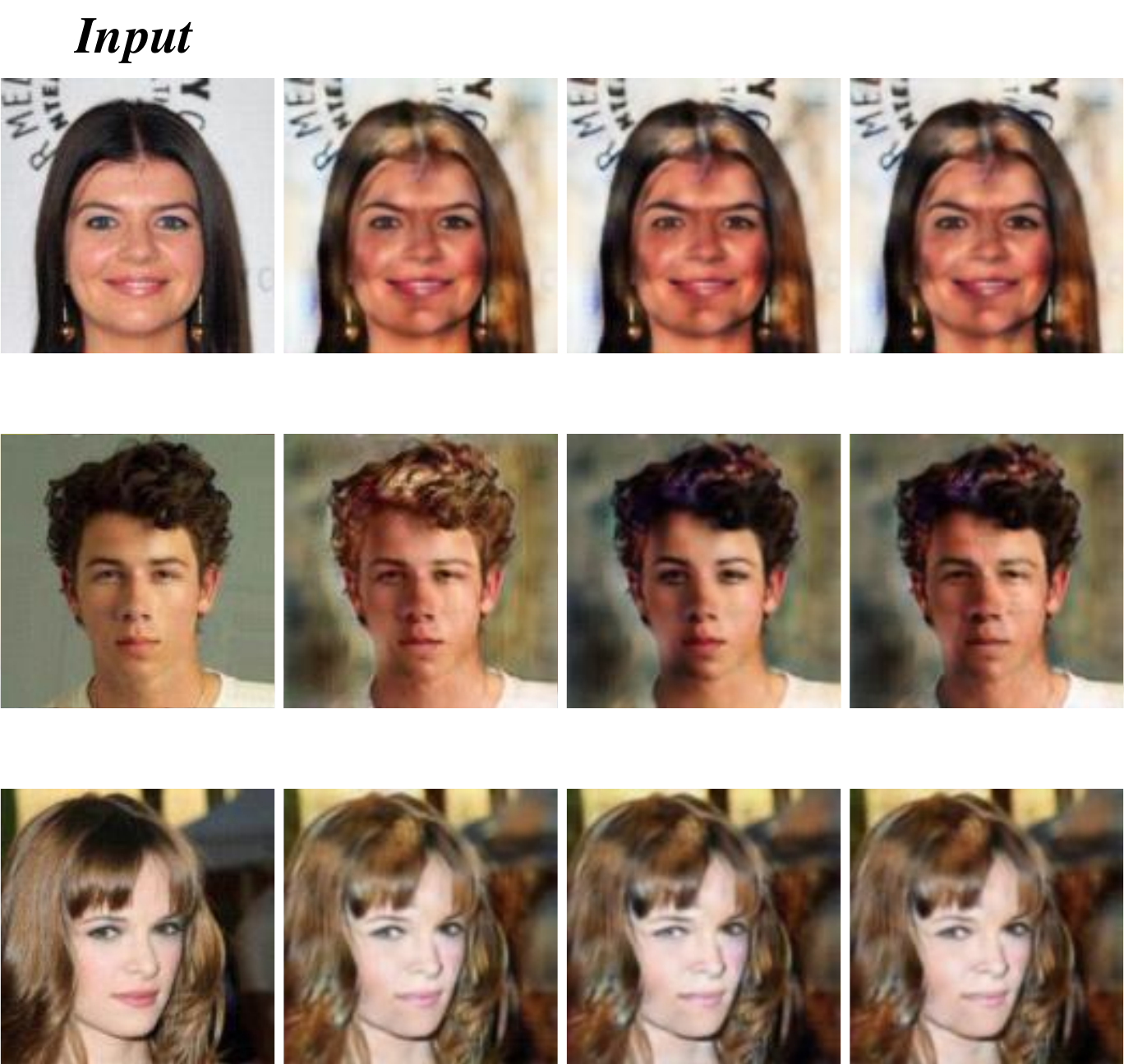}
		\end{minipage}%
	}
	\centering
	\caption{Exemplary results for StarGAN with the input of legitimate face images and adversarial face images. The generated adversarial example is applicable to arbitrary style translation (Hair color, Gender, and Aged).}
	\label{fig:6}
\end{figure*}

\begin{figure*}[ht!]
	\centering
	\subfloat[Original face translation]{
		\begin{minipage}[t]{0.48\linewidth}
			\centering
			\includegraphics[width=0.94\linewidth]{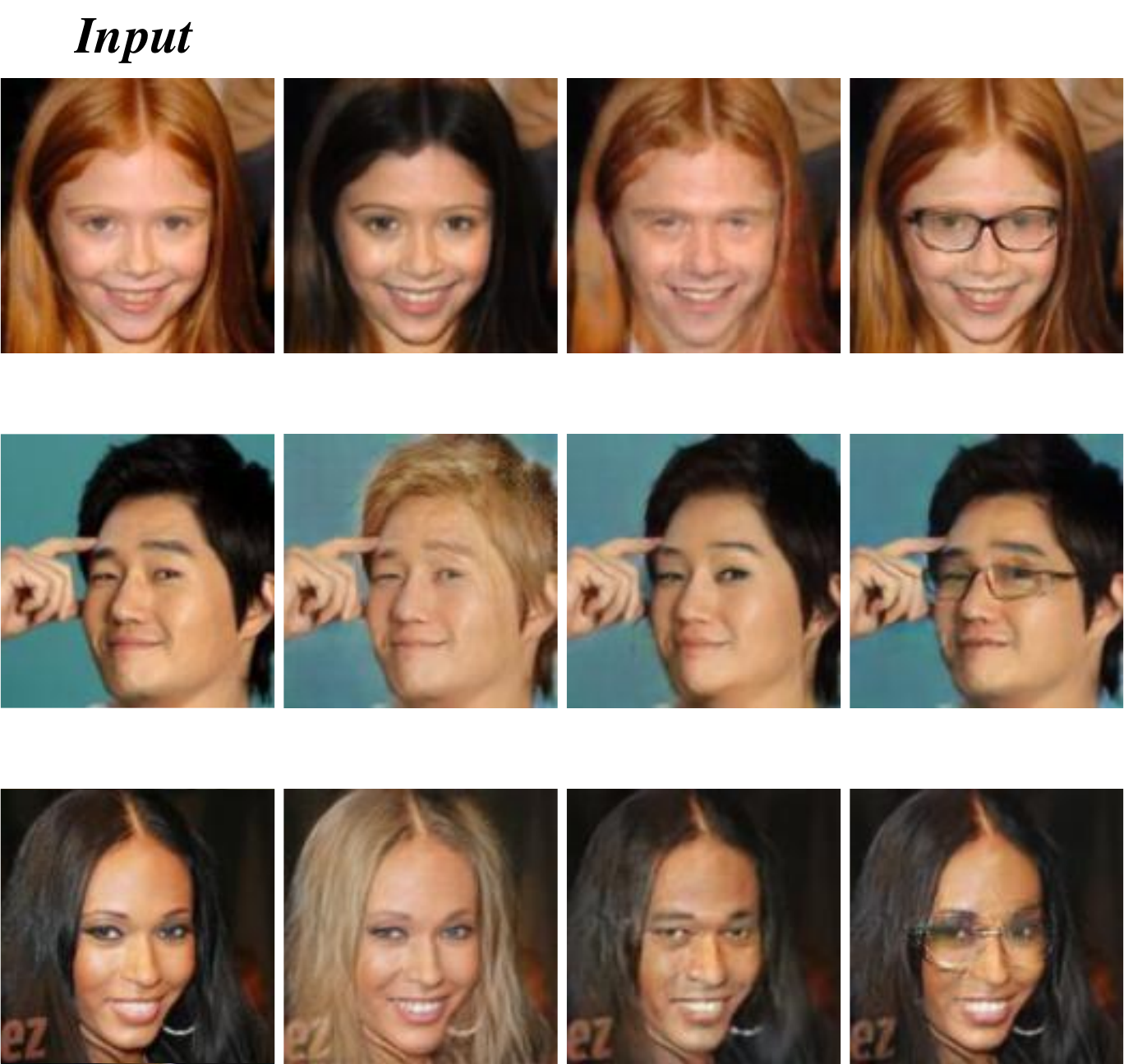}
		\end{minipage}%
	}
	\subfloat[Disrupted face translation]{
		\begin{minipage}[t]{0.48\linewidth}
			\centering
			\includegraphics[width=0.94\linewidth]{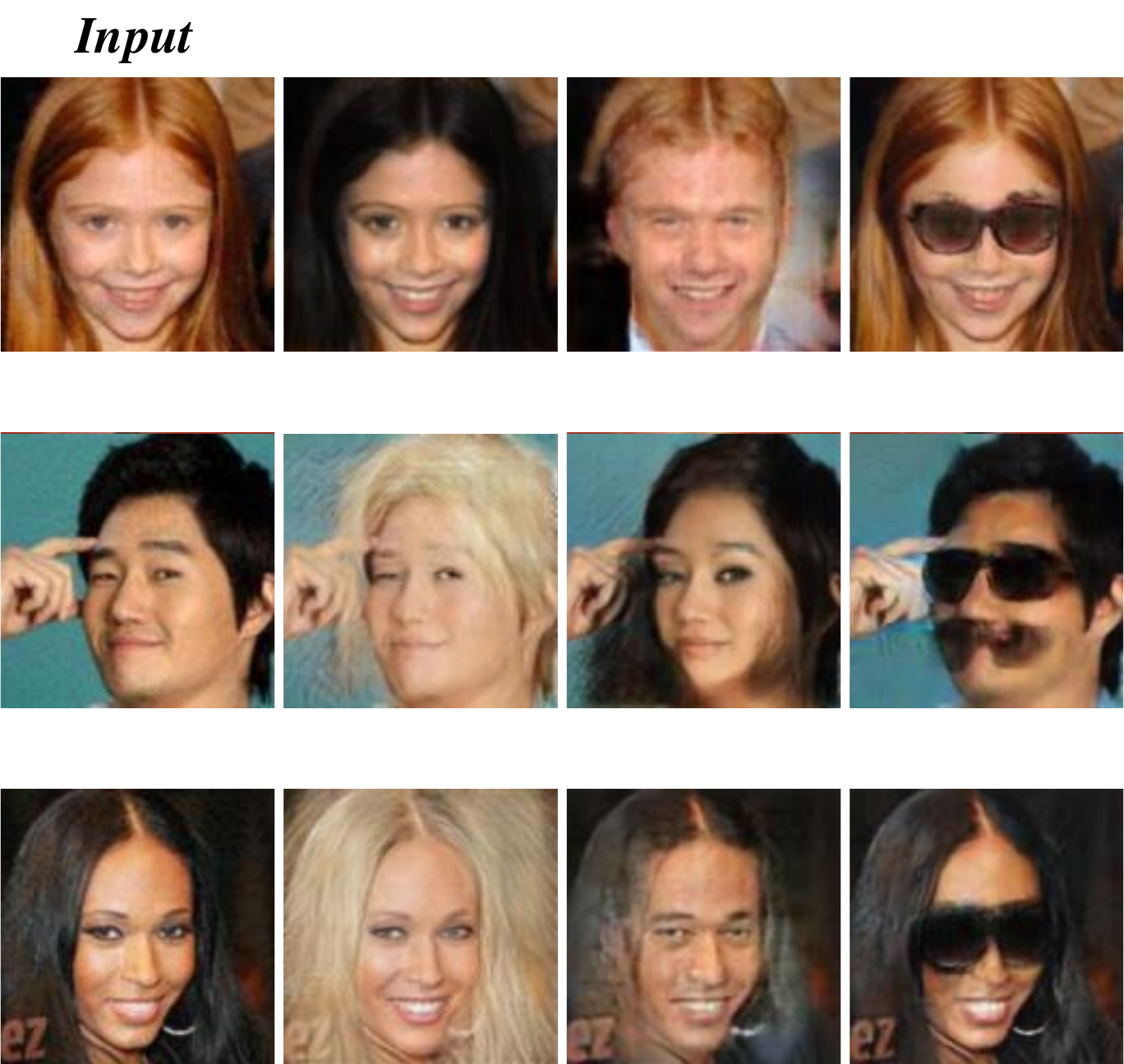}
		\end{minipage}%
	}
	\centering
	\caption{Exemplary results for AttGAN with the input of legitimate face images and adversarial face images. The generated adversarial example is applicable to arbitrary style translation (Hair color, Gender, and Glasses). }
	\label{fig:7}
\end{figure*}

\subsection{Ablation study}
In this section, we analyze the effectiveness of each component module in our proposed method. On account of only possessing the single field of legitimate examples, our cycle-consistent loss considers the unidirectional adversary circulation. The unidirectional generation maintains the consistency between legitimate examples and reconstructed images that are generated through two generators in TCA-GAN, respectively. Furthermore, the bidirectional cycle-consistent version of TCA-GAN considers the cycle reconstruction of adversarial examples on the basis of the unidirectional one. Besides, these adversarial examples against the substitute model are regenerated by PGD \cite{madry2017towards} per iteration. The comparison of the different versions of cycle-consistency is presented in Table \ref{tab:3}. It is worth mention that we mainly focus on perturbed face-swapped images. We can observe that TCA-GAN with the unidirectional cycle-consistency conducts much better than the one with bidirectional cycle-consistency. The main reason is that the obtained adversarial examples are excessively biased towards the substitute model, which is hard to transfer to other DNN models. On the contrary, our goal is to acquire a sub-optimal adversarial example and then generalize it to other face manipulation models like DeepFake. As a consequence, the guidance of an excessively powerful adversarial example may lead to an opposite effect to adversary transferring.

\begin{table}[!t]
	\centering
	\caption{Ablation results for component modules. The best result in each column is \textbf{bold}.
	}
	\renewcommand{\arraystretch}{1.4}
	\renewcommand{\tabcolsep}{2.1mm}
	\resizebox{0.99\linewidth}{!}{
		\begin{tabular}{c|ccc||c|c|c}
			\toprule
			&  {CC} & {LVD} & {PRM} & SSIM$\downarrow$ & FSIM$\downarrow$ & BRISQUE$\uparrow$ \\
			\midrule
			1 &  \checkmark & &  & 0.79 & 0.903 & 40.95 \\
			2 & & \checkmark & & 0.783 & 0.896 & 42.11  \\
			3 & & & \checkmark & 0.764 & 0.887 & 43.27  \\
			\cmidrule{1-7}
			4 & \checkmark & \checkmark & & 0.761 & 0.884 & 43.94  \\
			
			5 & & \checkmark & \checkmark  & 0.752 & 0.878 & 45.43\\
			
			6 &  \checkmark & & \checkmark & 0.755 & 0.882 & 45.18\\

			\cmidrule{1-7}
			7 &  \checkmark & \checkmark & \checkmark & \bf{0.731} & \bf{0.873} & \bf{47.13} \\
			\toprule
			\multicolumn{7}{l}{CC means the cycle consistency mechanism.}\\
			\multicolumn{7}{l}{LVD means the latent variable disruption.}\\
			\multicolumn{7}{l}{PRM means the post-regularization module.}
			\label{tab:4}
		\end{tabular}
	}
\end{table}

Concerning the main design of our proposed framework, we also discover the effect of component modules. Concretely, we mainly assess the disruption on face-swapped images in terms of the cycle-consistency, the latent variable disruption, and the post-regularization. The conducted ablation results are reported in Table \ref{tab:4}. Note that the latent variable disruption denotes that we induce a disruption on the latent variable of the substitute model when training TCA-GAN. As the table presented, the latent variable disruption shows a beneficial impression on enhancing the transferred adversary disruption. Moreover, we can derive a 3.1$\%$ SSIM performance enhancement with the latent variable disruption. The disruption on latent variables of the substitute model can be regarded as disturbing the facial feature during the face reconstruction. This also demonstrates that the perturbation towards the latent representation is more likely to generalize to other DNN models. Remarkably, the post-regularization presents a 3.9$\%$ SSIM performance boosting result. This post-processing can further help to distill the generated adversarial examples from TCA-GAN, meanwhile keeping a long discrepancy on the latent variables in terms of legitimate examples and regularized adversarial examples. Likewise, the result supports that the sub-optimal adversarial example has a more robust generalization performance than the optimal adversarial example. The proposed modules and the integration of them are demonstrated to be valid to disrupt the DeepFake face swapping mechanism. In summary, there is a large room for performance improvement on our proposed DeepFake face swapping dataset.

\begin{table}[t!]
	\centering
	\renewcommand{\arraystretch}{1.4}
	\renewcommand{\tabcolsep}{2.1mm}
	\caption{ The accuracy ($\%$) of several DeepFake detection methods on both original and distorted face-swapped images.}
	\resizebox{0.99\linewidth}{!}{
		\begin{threeparttable}

			\begin{tabular}{c||c|c}
				\toprule
				Method& Original accuracy & Distorted accuracy\cr
				\midrule		
				Xception \cite{rossler2019faceforensics++} &67.1\%
				&71.4\%\cr
				MesoNet \cite{afchar2018mesonet}&71.9\%
				&73.3\%\cr
				Capsule-forensics \cite{nguyen2019capsule}&60.5\%
				&65.4\%\cr
				EfficientNetB4Att\cite{bonettini2021video}&69.8\%
				&73.6\%\cr
				CNNDetection\cite{wang2020cnn}&78.3\%
				&79.7\%\cr
				
				\bottomrule
			\end{tabular}
			
			\label{tab:5}
		\end{threeparttable}
	}
\end{table}

\subsection{Enhancement on DeepFake detection}
Concurrently, DeepFake detection is also served as a solid mechanism to defend against DeepFake face swapping. Meanwhile, we explore several DeepFake detection methods to verify the detection facilitation of our methods. To be specific, we compare the detection success rate of original face-swapped images and disrupted face-swapped images. The comparison results of both original and distorted face-swapped image on our proposed dataset is presented in Table \ref{tab:5}. The table demonstrates that the distortion on face-swapped images can enhance the DeepFake detection performance. Moreover, it also shows the effectiveness of our DeepFake disruption method. The main reason behind the facilitation is that these DeepFake detection methods are focused on the imperceptible abnormal area of the input face images, while adversarial examples can enlarge the corresponding distortion of output face-swapped images. 

Note that under the disturbance of cross domain DeepFake examples, these DeepFake detection methods present a slightly poor performance on our DeepFake face swapping dataset. This also demonstrates the significance of our method to prevent malicious face swapping from the generation stage. Apart from distorting face-swapped images, our DeepFake disruption method can also be served as assistance for DeepFake detection.

\begin{table}[t!]
	\centering
	\renewcommand{\arraystretch}{1.4}
	\renewcommand{\tabcolsep}{2.1mm}
	\resizebox{0.99\linewidth}{!}{
		\begin{threeparttable}
			\caption{ Quantitative results of image quality assessment on the disrupted face style translation images.}
			
			\begin{tabular}{c||c|c|c}
				\toprule
				Method& SSIM$\downarrow$ & FSIM$\downarrow$ & BRISQUE$\uparrow$\cr
				\midrule		
				StarGAN \cite{choi2018stargan} &0.591
				&0.806&49.51\cr
				GANImation \cite{pumarola2018ganimation} &0.796
				&0.898&42.27\cr
				SaGAN \cite{zhang2018generative} &0.864
				&0.919&39.23\cr
				AttGAN \cite{he2019attgan} &0.841
				&0.908&43.76\cr
				
				\bottomrule
			\end{tabular}
			
			\label{tab:6}
		\end{threeparttable}
	}
\end{table}

\subsection{Generalization to face translation methods} 
In order to investigate the generalizability ability of our method on other face manipulation models, we transfer our adversarial attacks against several face image translation models in the restricted black-box scenario. With respect to StarGAN \cite{choi2018stargan} and AttGAN \cite{he2019attgan}, the corresponding face image translation results of both legitimate examples and adversarial examples are shown in Fig. \ref{fig:6} and Fig. \ref{fig:7}. It is clear that adversarial examples are nearly the same in human vision, yet they can strongly distort the face translation with respect different attributes. Note that different perturbed image translation results are all induced by the same input of adversarial example. In other words, the generated adversarial perturbation is image-agnostic against these face style translation models, thus can be adapted to arbitrary face images. 

To further quantity the visual distortion caused by restricted adversarial examples, we present image quality assessments on the disrupted face translation results as shown in Table \ref{tab:6}. Note that the face style translation is conducted on the CelebA \cite{liu2015deep} dataset. Moreover, we follow the same setting as the DeepFake disruption in the restricted black-box scenario. In consideration of the fixation of the adversarial perturbation upper bound, we mainly focus on the disturbance on face style-translated images. The generalization to face style translation models also represents that our proposed method can be served as an effective way to defend from face manipulation.


\section{Conclusion}
In this paper, we extensively explore the practical DeepFake defense scenario with a restricted black-box adversarial attack. We design the TCA-GAN method to generate transferrable adversarial perturbation via attacking a substitute model. Afterward, we adapt a novel post-regularization on the synthesized adversarial example to further enhance the generalization ability, which can induce intense disruption on DeepFake face swapping. For ease of subsequent researches on DeepFake disruption, we construct a DeepFake face swapping benchmark, including the face swapping models. To comprehensively evaluate the disruption induced by our adversarial examples, we conduct experiments on both referenced and non-referenced image quality assessments. Notably, we also show that our method can enhance the performance of DeepFake detection, which is more beneficial to defend against malicious face swapping. Subsequently, the extension to other face manipulation methods can further demonstrate the generalization performance of our method. We hope our work can shed light on protecting personal photos from unauthorized face manipulation in the real-world scenario.


%

%
%

\ifCLASSOPTIONcaptionsoff
  \newpage
\fi

\bibliographystyle{IEEEtran}
\bibliography{bare_jrnl}

\begin{thebibliography}{10}
\providecommand{\url}[1]{#1}
\csname url@samestyle\endcsname
\providecommand{\newblock}{\relax}
\providecommand{\bibinfo}[2]{#2}
\providecommand{\BIBentrySTDinterwordspacing}{\spaceskip=0pt\relax}
\providecommand{\BIBentryALTinterwordstretchfactor}{4}
\providecommand{\BIBentryALTinterwordspacing}{\spaceskip=\fontdimen2\font plus
\BIBentryALTinterwordstretchfactor\fontdimen3\font minus
  \fontdimen4\font\relax}
\providecommand{\BIBforeignlanguage}[2]{{%
\expandafter\ifx\csname l@#1\endcsname\relax
\typeout{** WARNING: IEEEtran.bst: No hyphenation pattern has been}%
\typeout{** loaded for the language `#1'. Using the pattern for}%
\typeout{** the default language instead.}%
\else
\language=\csname l@#1\endcsname
\fi
#2}}
\providecommand{\BIBdecl}{\relax}
\BIBdecl

\bibitem{korshunov2018deepfakes}
P.~Korshunov and S.~Marcel, ``Deepfakes: a new threat to face recognition?
  assessment and detection,'' \emph{arXiv preprint arXiv:1812.08685}, 2018.

\bibitem{rossler2019faceforensics++}
A.~Rossler, D.~Cozzolino, L.~Verdoliva, C.~Riess, J.~Thies, and M.~Nie{\ss}ner,
  ``Faceforensics++: Learning to detect manipulated facial images,'' in
  \emph{Proceedings of the IEEE/CVF International Conference on Computer
  Vision}, 2019, pp. 1--11.

\bibitem{afchar2018mesonet}
D.~Afchar, V.~Nozick, J.~Yamagishi, and I.~Echizen, ``Mesonet: a compact facial
  video forgery detection network,'' in \emph{2018 IEEE International Workshop
  on Information Forensics and Security (WIFS)}.\hskip 1em plus 0.5em minus
  0.4em\relax IEEE, 2018, pp. 1--7.

\bibitem{nguyen2019capsule}
H.~H. Nguyen, J.~Yamagishi, and I.~Echizen, ``Capsule-forensics: Using capsule
  networks to detect forged images and videos,'' in \emph{ICASSP 2019-2019 IEEE
  International Conference on Acoustics, Speech and Signal Processing
  (ICASSP)}.\hskip 1em plus 0.5em minus 0.4em\relax IEEE, 2019, pp. 2307--2311.

\bibitem{He_2016_CVPR}
K.~He, X.~Zhang, S.~Ren, and J.~Sun, ``Deep residual learning for image
  recognition,'' in \emph{Proceedings of the IEEE Conference on Computer Vision
  and Pattern Recognition (CVPR)}, June 2016.

\bibitem{ronneberger2015u}
O.~Ronneberger, P.~Fischer, and T.~Brox, ``U-net: Convolutional networks for
  biomedical image segmentation,'' in \emph{International Conference on Medical
  image computing and computer-assisted intervention}.\hskip 1em plus 0.5em
  minus 0.4em\relax Springer, 2015, pp. 234--241.

\bibitem{NIPS2014_5423}
I.~Goodfellow, J.~Pouget-Abadie, M.~Mirza, B.~Xu, D.~Warde-Farley, S.~Ozair,
  A.~Courville, and Y.~Bengio, ``Generative adversarial nets,'' in \emph{NIPS},
  2014, pp. 2672--2680.

\bibitem{journals/corr/SzegedyZSBEGF13}
C.~Szegedy, W.~Zaremba, I.~Sutskever, J.~Bruna, D.~Erhan, I.~Goodfellow, and
  R.~Fergus, ``Intriguing properties of neural networks,'' in \emph{ICLR},
  2014.

\bibitem{9325048}
Z.~Che, A.~Borji, G.~Zhai, S.~Ling, J.~Li, Y.~Tian, G.~Guo, and P.~Le~Callet,
  ``Adversarial attack against deep saliency models powered by non-redundant
  priors,'' \emph{IEEE Transactions on Image Processing}, vol.~30, pp.
  1973--1988, 2021.

\bibitem{9430730}
L.~Gao, Z.~Huang, J.~Song, Y.~Yang, and H.~T. Shen, ``Push pull: Transferable
  adversarial examples with attentive attack,'' \emph{IEEE Transactions on
  Multimedia}, pp. 1--1, 2021.

\bibitem{kos2018adversarial}
J.~Kos, I.~Fischer, and D.~Song, ``Adversarial examples for generative
  models,'' in \emph{SPW}, 2018, pp. 36--42.

\bibitem{DBLP:journals/corr/GoodfellowSS14}
I.~Goodfellow, J.~Shlens, and C.~Szegedy, ``Explaining and harnessing
  adversarial examples,'' in \emph{ICLR}, 2015.

\bibitem{xiao2018generating}
C.~Xiao, B.~Li, J.-Y. Zhu, W.~He, M.~Liu, and D.~Song, ``Generating adversarial
  examples with adversarial networks,'' in \emph{Proceedings of the 27th
  International Joint Conference on Artificial Intelligence}, 2018, pp.
  3905--3911.

\bibitem{chen2017zoo}
P.-Y. Chen, H.~Zhang, Y.~Sharma, J.~Yi, and C.-J. Hsieh, ``Zoo: Zeroth order
  optimization based black-box attacks to deep neural networks without training
  substitute models,'' in \emph{Proceedings of the 10th ACM workshop on
  artificial intelligence and security}, 2017, pp. 15--26.

\bibitem{blanz2004exchanging}
V.~Blanz, K.~Scherbaum, T.~Vetter, and H.-P. Seidel, ``Exchanging faces in
  images,'' in \emph{Computer Graphics Forum}, 2004, pp. 669--676.

\bibitem{bitouk2008face}
D.~Bitouk, N.~Kumar, S.~Dhillon, P.~Belhumeur, and S.~K. Nayar, ``Face
  swapping: automatically replacing faces in photographs,'' in \emph{ACM
  SIGGRAPH}, 2008, pp. 1--8.

\bibitem{8237659}
I.~{Korshunova}, W.~{Shi}, J.~{Dambre}, and L.~{Theis}, ``Fast face-swap using
  convolutional neural networks,'' in \emph{ICCV}, 2017, pp. 3697--3705.

\bibitem{bao2018towards}
J.~Bao, D.~Chen, F.~Wen, H.~Li, and G.~Hua, ``Towards open-set identity
  preserving face synthesis,'' in \emph{Proceedings of the IEEE Conference on
  Computer Vision and Pattern Recognition}, 2018, pp. 6713--6722.

\bibitem{li2019faceshifter}
L.~Li, J.~Bao, H.~Yang, D.~Chen, and F.~Wen, ``Faceshifter: Towards high
  fidelity and occlusion aware face swapping,'' \emph{arXiv preprint
  arXiv:1912.13457}, 2019.

\bibitem{nirkin2019fsgan}
Y.~Nirkin, Y.~Keller, and T.~Hassner, ``Fsgan: Subject agnostic face swapping
  and reenactment,'' in \emph{Proceedings of the IEEE/CVF International
  Conference on Computer Vision}, 2019, pp. 7184--7193.

\bibitem{chen2020simswap}
R.~Chen, X.~Chen, B.~Ni, and Y.~Ge, ``Simswap: An efficient framework for high
  fidelity face swapping,'' in \emph{Proceedings of the 28th ACM International
  Conference on Multimedia}, 2020, pp. 2003--2011.

\bibitem{kurakin2016adversarial}
A.~Kurakin, I.~Goodfellow, and S.~Bengio, ``Adversarial examples in the
  physical world,'' \emph{arXiv preprint arXiv:1607.02533}, 2016.

\bibitem{madry2017towards}
A.~Madry, A.~Makelov, L.~Schmidt, D.~Tsipras, and A.~Vladu, ``Towards deep
  learning models resistant to adversarial attacks,'' \emph{arXiv preprint
  arXiv:1706.06083}, 2017.

\bibitem{moosavi2017universal}
S.-M. Moosavi-Dezfooli, A.~Fawzi, O.~Fawzi, and P.~Frossard, ``Universal
  adversarial perturbations,'' in \emph{CVPR}, 2017, pp. 1765--1773.

\bibitem{8601309}
J.~{Su}, D.~V. {Vargas}, and K.~{Sakurai}, ``One pixel attack for fooling deep
  neural networks,'' \emph{IEEE Transactions on Evolutionary Computation},
  vol.~23, no.~5, pp. 828--841, 2019.

\bibitem{croce2020minimally}
F.~Croce and M.~Hein, ``Minimally distorted adversarial examples with a fast
  adaptive boundary attack,'' in \emph{International Conference on Machine
  Learning}.\hskip 1em plus 0.5em minus 0.4em\relax PMLR, 2020, pp. 2196--2205.

\bibitem{9186644}
H.~Zhang, Y.~Avrithis, T.~Furon, and L.~Amsaleg, ``Walking on the edge: Fast,
  low-distortion adversarial examples,'' \emph{IEEE Transactions on Information
  Forensics and Security}, vol.~16, pp. 701--713, 2021.

\bibitem{9252132}
Y.~Zhong and W.~Deng, ``Towards transferable adversarial attack against deep
  face recognition,'' \emph{IEEE Transactions on Information Forensics and
  Security}, vol.~16, pp. 1452--1466, 2021.

\bibitem{tabacof2016adversarial}
P.~Tabacof, J.~Tavares, and E.~Valle, ``Adversarial images for variational
  autoencoders,'' \emph{arXiv preprint arXiv:1612.00155}, 2016.

\bibitem{ruiz2020disrupting}
N.~Ruiz and S.~Sclaroff, ``Disrupting deepfakes: Adversarial attacks against
  conditional image translation networks and facial manipulation systems,''
  \emph{arXiv preprint arXiv:2003.01279}, 2020.

\bibitem{ruiz2020protecting}
N.~Ruiz, S.~A. Bargal, and S.~Sclaroff, ``Protecting against image translation
  deepfakes by leaking universal perturbations from black-box neural
  networks,'' \emph{arXiv preprint arXiv:2006.06493}, 2020.

\bibitem{fang2020adversarial}
Z.~Fang, Y.~Yang, J.~Lin, and R.~Zhan, ``Adversarial attacks for multi target
  image translation networks,'' in \emph{2020 IEEE International Conference on
  Progress in Informatics and Computing (PIC)}.\hskip 1em plus 0.5em minus
  0.4em\relax IEEE, 2020, pp. 179--184.

\bibitem{Dong_deepfake}
J.~Dong and X.~Xie, ``Visually maintained image disturbance against deepfake
  face swapping,'' in \emph{2021 IEEE International Conference on Multimedia
  and Expo (ICME)}, 2021.

\bibitem{sun2020landmark}
P.~Sun, Y.~Li, H.~Qi, and S.~Lyu, ``Landmark breaker: Obstructing deepfake by
  disturbing landmark extraction,'' in \emph{2020 IEEE International Workshop
  on Information Forensics and Security (WIFS)}.\hskip 1em plus 0.5em minus
  0.4em\relax IEEE, 2020, pp. 1--6.

\bibitem{papernot2017practical}
N.~Papernot, P.~McDaniel, I.~Goodfellow, S.~Jha, Z.~B. Celik, and A.~Swami,
  ``Practical black-box attacks against machine learning,'' in
  \emph{Proceedings of the 2017 ACM on Asia conference on computer and
  communications security}, 2017, pp. 506--519.

\bibitem{inkawhich2019feature}
N.~Inkawhich, W.~Wen, H.~H. Li, and Y.~Chen, ``Feature space perturbations
  yield more transferable adversarial examples,'' in \emph{Proceedings of the
  IEEE/CVF Conference on Computer Vision and Pattern Recognition}, 2019, pp.
  7066--7074.

\bibitem{huang2019enhancing}
Q.~Huang, I.~Katsman, H.~He, Z.~Gu, S.~Belongie, and S.-N. Lim, ``Enhancing
  adversarial example transferability with an intermediate level attack,'' in
  \emph{Proceedings of the IEEE/CVF International Conference on Computer
  Vision}, 2019, pp. 4733--4742.

\bibitem{liu2015deep}
Z.~Liu, P.~Luo, X.~Wang, and X.~Tang, ``Deep learning face attributes in the
  wild,'' in \emph{Proceedings of the IEEE international conference on computer
  vision}, 2015, pp. 3730--3738.

\bibitem{wang2004image}
Z.~Wang, A.~C. Bovik, H.~R. Sheikh, and E.~P. Simoncelli, ``Image quality
  assessment: from error visibility to structural similarity,'' \emph{IEEE
  transactions on image processing}, vol.~13, no.~4, pp. 600--612, 2004.

\bibitem{zhang2011fsim}
L.~Zhang, L.~Zhang, X.~Mou, and D.~Zhang, ``Fsim: A feature similarity index
  for image quality assessment,'' \emph{IEEE transactions on Image Processing},
  vol.~20, no.~8, pp. 2378--2386, 2011.

\bibitem{mittal2012no}
A.~Mittal, A.~K. Moorthy, and A.~C. Bovik, ``No-reference image quality
  assessment in the spatial domain,'' \emph{IEEE Transactions on image
  processing}, vol.~21, no.~12, pp. 4695--4708, 2012.

\bibitem{xie2019improving}
C.~Xie, Z.~Zhang, Y.~Zhou, S.~Bai, J.~Wang, Z.~Ren, and A.~L. Yuille,
  ``Improving transferability of adversarial examples with input diversity,''
  in \emph{Proceedings of the IEEE/CVF Conference on Computer Vision and
  Pattern Recognition}, 2019, pp. 2730--2739.

\bibitem{dong2019evading}
Y.~Dong, T.~Pang, H.~Su, and J.~Zhu, ``Evading defenses to transferable
  adversarial examples by translation-invariant attacks,'' in \emph{Proceedings
  of the IEEE/CVF Conference on Computer Vision and Pattern Recognition}, 2019,
  pp. 4312--4321.

\bibitem{croce2020reliable}
F.~Croce and M.~Hein, ``Reliable evaluation of adversarial robustness with an
  ensemble of diverse parameter-free attacks,'' in \emph{International
  Conference on Machine Learning}.\hskip 1em plus 0.5em minus 0.4em\relax PMLR,
  2020, pp. 2206--2216.

\bibitem{bonettini2021video}
N.~Bonettini, E.~D. Cannas, S.~Mandelli, L.~Bondi, P.~Bestagini, and S.~Tubaro,
  ``Video face manipulation detection through ensemble of cnns,'' in \emph{2020
  25th International Conference on Pattern Recognition (ICPR)}.\hskip 1em plus
  0.5em minus 0.4em\relax IEEE, 2021, pp. 5012--5019.

\bibitem{wang2020cnn}
S.-Y. Wang, O.~Wang, R.~Zhang, A.~Owens, and A.~A. Efros, ``Cnn-generated
  images are surprisingly easy to spot... for now,'' in \emph{Proceedings of
  the IEEE/CVF Conference on Computer Vision and Pattern Recognition}, 2020,
  pp. 8695--8704.

\bibitem{choi2018stargan}
Y.~Choi, M.~Choi, M.~Kim, J.-W. Ha, S.~Kim, and J.~Choo, ``Stargan: Unified
  generative adversarial networks for multi-domain image-to-image
  translation,'' in \emph{Proceedings of the IEEE conference on computer vision
  and pattern recognition}, 2018, pp. 8789--8797.

\bibitem{pumarola2018ganimation}
A.~Pumarola, A.~Agudo, A.~M. Martinez, A.~Sanfeliu, and F.~Moreno-Noguer,
  ``Ganimation: Anatomically-aware facial animation from a single image,'' in
  \emph{Proceedings of the European conference on computer vision (ECCV)},
  2018, pp. 818--833.

\bibitem{zhang2018generative}
G.~Zhang, M.~Kan, S.~Shan, and X.~Chen, ``Generative adversarial network with
  spatial attention for face attribute editing,'' in \emph{Proceedings of the
  European conference on computer vision (ECCV)}, 2018, pp. 417--432.

\bibitem{he2019attgan}
Z.~He, W.~Zuo, M.~Kan, S.~Shan, and X.~Chen, ``Attgan: Facial attribute editing
  by only changing what you want,'' \emph{IEEE Transactions on Image
  Processing}, vol.~28, no.~11, pp. 5464--5478, 2019.

\end{thebibliography}


\end{document}